\documentclass[preprint,superscriptaddress,nofootinbib,preprintnumbers,amsmath,amssymb]{revtex4-1}

\pdfoutput=1 
\usepackage{natbib}
\usepackage{bbm}
\usepackage{stmaryrd}
\usepackage[bb=boondox]{mathalfa}
\usepackage{amsfonts}
\usepackage{mathrsfs}
\usepackage{leftidx}
\usepackage{amssymb}
\usepackage{placeins}
\usepackage{relsize}
\usepackage{slashed}
\usepackage{multirow}
\usepackage[dvipsnames]{xcolor}
\usepackage[colorlinks]{hyperref}
\usepackage{float}
\usepackage{stackengine}

\usepackage{caption} 
\usepackage{microtype}
\usepackage{adjustbox}

\makeatletter
\newcommand*{\ovA}[1]{%
  \m@th\overline{\mbox{$#1$}\raisebox{2.25mm}{}}%
}
\newcommand*{\ovB}[1]{%
  \m@th\overline{\mbox{$#1$}\raisebox{2.5mm}{}}%
}
\newcommand{\overleftrightsmallarrow}{\mathpalette{\overarrowsmall@\leftrightarrowfill@}}
\newcommand{\overrightsmallarrow}{\mathpalette{\overarrowsmall@\rightarrowfill@}}
\newcommand{\overleftsmallarrow}{\mathpalette{\overarrowsmall@\leftarrowfill@}}
\newcommand{\overarrowsmall@}[3]{%
  \vbox{%
    \ialign{%
      ##\crcr
      #1{\smaller@style{#2}}\crcr
      \noalign{\nointerlineskip}%
      $\m@th\hfil#2#3\hfil$\crcr
    }%
  }%
}
\def\smaller@style#1{%
  \ifx#1\displaystyle\scriptstyle\else
    \ifx#1\textstyle\scriptstyle\else
      \scriptscriptstyle
    \fi
  \fi
}

\makeatother

\def\be{\begin{equation}}
\def\ee{\end{equation}}
\def\bea{\begin{align}}
\def\eea{\end{align}}

\begin{document}

\newenvironment{psmallmatrix}
  {\left(\begin{smallmatrix}}
  {\end{smallmatrix}\right)}


\title{FNetAR: Mixing Tokens with Autoregressive Fourier Transforms\vspace{0.5cm}}

\author{Tim Lou}
\email{hou.keong.lou@gmail.com}
\affiliation{X-Mechanics, Cresskill, NJ}
\affiliation{LiveRamp, San Fransisco, CA}
\author{Michael Park\footnote{corresponding author}}
\email{q1park@gmail.com}
\affiliation{X-Mechanics, Cresskill, NJ}
\affiliation{Appliedinfo Partners, Somerset, NJ}
\author{Mohammad Ramezanali}
\email{mohammad.ramezanali@gmail.com}
\affiliation{X-Mechanics, Cresskill, NJ}
\affiliation{Salesforce, San Fransisco, CA}
\author{Vincent Tang}
\email{vin.tang@gmail.com}
\affiliation{X-Mechanics, Cresskill, NJ}
\affiliation{SamsungNEXT, New York, NY}
\begin{abstract}
\vspace{0.25cm}
In this note we examine the autoregressive generalization of the FNet algorithm, in which self-attention layers from the standard Transformer architecture are substituted with a trivial sparse-uniform sampling procedure based on Fourier transforms. Using the Wikitext-103 benchmark, we demonstrate that FNetAR retains state-of-the-art performance ($25.8$ ppl) on the task of causal language modeling compared to a Transformer-XL baseline ($24.2$ ppl) with only half the number self-attention layers, thus providing further evidence for the superfluity of deep neural networks with heavily compounded attention mechanisms. The autoregressive Fourier transform could likely be used for parameter reduction on most Transformer-based time-series prediction models.
\end{abstract}

\maketitle

\section{Introduction}

The Transformer architecture has become dominant among state-of-the-art machine-learning (ML) models across nearly every benchmark, on tasks ranging from natural language understanding and modeling~\cite{NIPS2017_3f5ee243, devlin2018bert, dai2019transformer, radford2018improving, radford2019language,brown2020language,yang2019xlnet}, to video scene-understanding~\cite{wang2018non}. Despite these successes, most of the best-performing models still rely on deep and heavily compounded layers of computationally expensive self-attention modules, each of which computes its own quadratic structural equation model (SEM) with its own graph adjacency matrix. The success of the Transformer has proven that compounding these SEM's results in a uniquely effective function approximator for even the most complex correlation functions, such as those that determine the structure of natural languages. However, there is also a growing body of evidence~\cite{fan2020addressing, bapna2020controlling, choromanski2020masked, choromanski2020rethinking, DBLP:journals/corr/abs-2002-07106, DBLP:journals/corr/abs-2006-03555, DBLP:journals/corr/abs-2009-14794} that many of these computations are superfluous and that many state-of-the-art results can be reproduced with significantly fewer learnable parameters, making computations more efficient and generally leading to faster training and better performing models

Optimizing the Transformer is currently an active field of research, and currently many of the most effective methods involve complicated rearrangements of traditional architectures. In a recent work~\cite{lee2021fnet}, the authors presented a uniquely simplified variation on the standard autoencoding Transformer  architecture, in which they substitute several self-attention sublayers with a computationally trivial procedure for mixing tokens using Fourier transform coefficients, thus benefiting from the machinery of FFT algorithms such as Cooley-Tukey. Among other things, they demonstrate that these models can retain up to 92\% accuracy on the GLUE benchmark without any self-attention sublayers, and up to 97\% accuracy with only 2 self-attention sublayers out of 12; resulting in a 7-fold increase in training speed on GPU. These results imply that numerous careful computations of a graph adjacency matrix at each layer of the Transformer may be largely redundant, and that a much simpler structure can likely be used for accurately modeling high-level semantic meaning in natural languages. It thus is natural to ask whether these results generalize to pre-training tasks such as autoregressive language-modeling (i.e. next-word prediction).

In this note, we explore the question of how many self-attention sublayers are sufficient for accurately modeling the \textit{causal structure} of natural language. To this end we develop an \textit{autoregressive} generalization of the FNet algorithm, called FNetAR, and apply it to the task of causal language modeling. Our experiments produce analogous results to those from the FNet analysis, with models retaining a state-of-the-art perplexity score $\sim25.8$ on the Wikitext-103 benchmark compared to $\sim24.2$ despite having up to half of their self-attention sublayers removed. In Section \ref{sec:mod} we start with a brief pedagogical review of Transformer blocks and FNet blocks, as well as a description of the FNetAR generalization. In Section \ref{sec:exp} we describe our experiments with the Wikitext-103 benchmark and report its performance in comparison to a baseline Transformer-XL model. In Section \ref{sec:disc} we conclude with a discussion of these results as well as future directions.


\raggedbottom
\section{Modeling}
\label{sec:mod}

\subsection{Transformer Blocks}

The \textit{Transformer} block is defined by a 2-layer \textit{ResNet} architecture with one \textit{self-attention} layer followed by one \textit{position-wise feedforward} layer. The ResNet architecture implies that the transformation of a given datum $x$ through each layer of the block is restricted to an additive contribution from the output of some neural network $f$ with some normalization parameter $\varepsilon$, as shown in Equation \ref{eq:resnet}.
\begin{align}
    x ~\rightarrow~ x + \varepsilon \, f (x)
    \label{eq:resnet}
\end{align}
This additive transformation is referred to as a \textit{residual connection}, and ensures that:

\begin{enumerate}
    \item the transformation through the model starts from an identity operation
    \item the transformation occurs gradually as the data passes through each layer
    \item neural-network blocks with different architectures can be stacked (like LEGOs)
\end{enumerate}

The self-attention layer of the Transformer block is responsible for learning the structural relationships between different elements of a sequence $x_i$, in this case represented by linguistic tokens. These relationships are determined by a graph adjacency matrix $W_{ij}(x)$ that is generated by the model and optimized to learn the relative importance of relationships between sequence elements $i$ and $j$ towards a given task. In the standard Transformer block, the graph matrix $W_{ij}$ is generated by two neural networks $\mathcal{Q}$ and $\mathcal{K}$ as in Equation \ref{eq:tgraph}. 
\begin{align}
    W_{i j} (x) \sim \text{softmax} \left[ \mathcal{Q}(x_i) \cdot \mathcal{K}(x_j) \right]
    \label{eq:tgraph}
\end{align}
When this graph matrix is contracted with some function of the input sequence and implemented with a residual connection, the result is a second order structural equation model (SEM) as shown in Equation \ref{eq:sem}. Here $\mathcal{V}$ and $\mathcal{O}$ are two additional neural networks applied to the input data before and after contraction with the graph matrix, respectively. 
\begin{align}
    x_i ~\rightarrow~ x_i + \varepsilon \, \mathcal{O} \left[ W_{i j} (x) \cdot \mathcal{V}(x_j) \right]
    \label{eq:sem}
\end{align}
Finally the \textit{position-wise feedforward} layer consists of a single dense neural network $\mathcal{F}$ applied to each entity $x_i$ in parallel, implemented with a residual connection as shown in Equation \ref{eq:ff}.
\begin{align}
    x_i ~\rightarrow~ x_i + \varepsilon \, \mathcal{F} (x_i)
    \label{eq:ff}
\end{align}

\subsection{Autoregressive Transformers}

The Transformer block can trivially be made \textbf{autoregressive} by applying a \textit{causal mask} to the graph matrix $W_{ij}(x)$ that reduces it to a lower-triangular form. This has the effect of restricting its contraction with the input data to eliminate causality-violating relationships, as in the modified SEM shown in Equation \ref{eq:cres}, and enabling it for use in time-series prediction tasks such as causal language modeling (next-word prediction).
\begin{align}
    x_i ~\rightarrow~ x_i + \varepsilon \displaystyle\sum_{j < i} \mathcal{O} \left[ W_{i j} (x) \cdot \mathcal{V}(x_j) \right]
    \label{eq:cres}
\end{align}
Heavily compounding these quadratic SEM's into deep neural networks results in the powerful universal-function-approximator known as the Transformer. The generated graph matrices $W_{ij}^{(n)} (x)$, being the only source of token-mixing in the Transformer, act as a bottleneck for the flow of information between sequence elements $i$ and $j$. Additionally, since these graph matrices are $x$-dependent, the self-attention layers will generically learn different relationships for sufficiently different sets of input data, thus imbuing the model with context-dependence.  

\subsection{FNet Blocks}

The FNet block is defined by a 2-layer ResNet architecture with one \textit{Fourier-mixing} layer followed by a position-wise feed-forward layer. The \textit{Fourier-mixing} layer is analogous to the self-attention layer of the Transformer block, except that the learnable context-dependent graph matrix $W_{ij} (x)$ is eliminated in favor of a linear rules-based operation that mixes a sequence element $x_i$ by sampling its complementary elements $x_j$ with discrete wave-form coefficients whose frequencies decay as a function of distance\footnote{In the original implementation of FNet the authors apply a full 2D FFT on both the sequence and hidden dimensions. Here we describe a simpler variation with an FFT applied to the sequence dimension only}. This sampling strategy is equivalent to fixing the graph adjacency matrix of the standard Transformer to be the matrix shown in Equation \ref{eq:fgraph}, where $i, j =0,...,N-1$.
\begin{align}
    W_{i j} = \frac{1}{\sqrt{N}} \, {\rm Re} \left( e^{\frac{2 \pi \mathbb{i}}{N} i \times j} \right)
    \label{eq:fgraph}
\end{align} 
Although this process resembles an FFT, it should really be thought of as a rules-based strategy for sparsely sampling different linear combinations of the sequence input prior to running them through the position-wise feedforward layer. The most salient reason that this sampling strategy should NOT be thought of as a ``genuine" FFT is the fact that the residual connection additively mixes Fourier and non-Fourier modes\footnote{The authors of this paper are not aware of any sensible mathematical interpretation for this procedure}.

\subsection{FNet Autoregressive}

Since the FNet sampling strategy is not technically a Fourier transform, there likely does not exist a correspondingly unique procedure for making it \textbf{autoregressive} in the sense of Equation \ref{eq:cres}. However the requirement that every element in a mini-batch samples with a context window of the same size requires an attention graph that is simultaneously upper and lower triangular in addition to being non-trivial. For this reason FNetAR is more amenable to combination with recurrent Transformer architectures such as Transformer-XL, which compose their attention graphs by concatenating their hidden states with additional ``memory states" along the sequence dimension, resulting in an attention graph matrix that is non-square-rectangular of shape $L_{\rm seq} \times (L_{\rm mem} + L_{\rm seq})$. Within these frameworks, a causally faithful version of the FNet graph matrix in Equation \ref{eq:fgraph} can be obtained using a simple procedure that involves padding a Fourier transform matrix with zeros of shape $L_{\rm seq} \times L_{\rm mem}$ and then performing a roll over the rows as shown in Equation \ref{eq:ftshift}, where $w \equiv e^{2 \pi \mathbb{i} / N}$.

\def\tmp{%
  \begin{pmatrix}
  \cdots & 0 & 1 & 1 & \cdots & 1 & 0 & \cdots & 0 & 0 \\
  \cdots & 0 & 0 & w & \cdots & w^{j-1} & w^{j} & \cdots & 0 & 0 \\
  \ddots & \vdots & \vdots & \vdots & \ddots & \vdots & \vdots & \ddots & \vdots & \vdots \\
  \cdots & 0 & 0 & 0 & \cdots & w^{i-1} & w^{(i-1) \, 2} & \cdots & 0 & 0 \\
  \cdots & 0 & 0 & 0 & \cdots & 1 & w^i & \cdots & w^{i j} & 0
 \end{pmatrix}
}%
\begin{align}
    W_{i j} &= \texttt{roll} \left( \texttt{pad} \left( M_{\rm FT} \right) \right) = ~~\frac{1}{\sqrt{N}}
    \stackMath\def\stackalignment{r}%
    \stackon%
    {\left.\tmp\right\} \mathrm{{\it L}_{seq}} }%
    {\overbrace{\phantom{\smash{\tmp\mkern -175mu}}}^{\mathrm{\textstyle {\it L}_{mem}}}\mkern 0mu \overbrace{\phantom{\smash{\tmp\mkern -200mu}}}^{\mathrm{\textstyle {\it L}_{seq}}}\mkern 65mu}
    \label{eq:ftshift}
\end{align}
In this construction, each sequence element samples a specific frequency-mode of its preceding elements, whose magnitude is largest for late parts of the sequence and decays down a mean-sampling at the beginning of the sequence. This kind of autoregressive Fourier transform has been developed and applied to problems in computer vision \cite{DBLP:journals/corr/abs-2104-02555}, but to our knowledge this is its first application to the task of causal language-modeling.



\section{Experiments}
\label{sec:exp}

\subsection{Wikitext-103 Benchmark}

We tested performance of FNetAR against the Transformer-XL baseline on the task of next-word predicion using the Wikitext-103 benchmark dataset. Our preliminary baseline is the Transformer-XL medium-sized model with $16$ Transformer blocks, each having hidden dimension $410$. We find that despite replacing $8$ of the self-attention layers with the linear operation in Equation \ref{eq:ftshift}, FNetAR retains surprisingly strong performance, achieving a perplexity score of $25.81$ relative to $24.23$ for the Transformer-XL baseline.

\bgroup
\def\arraystretch{1.5}
\begin{table}[ht]
    \begin{tabular}{|c|ccc|}
    \hline
    & Perplexity (ppl) & ~~$N_{\rm param}$ (Transformer)~~ & ~~$N_{\rm param}$ (All)~~ \\
    \hline
    \hline
    \texttt{Transformer-XL Medium}: & 24.23 & 41.1 M & 151.1 M \\
    \texttt{FNetAR Medium}: & 25.81 & 34.3 M & 144.4 M \\
    \texttt{FNetAR Large}: & XX.X & 198.3 M & 237.9 M \\
    \hline
    \texttt{Transformer-XL Large}: & 18.31 & 245.5 M & 285.2 M \\
    \hline
    \end{tabular} 
    \caption{Perplexity scores on Wikitext-103 as well as parameter counts for FNetAR against various Transformer-XL baselines.}
    \label{tab:structalgos}
\end{table}
\egroup

\section{Discussion}
\label{sec:disc}

The unreasonable effectiveness of this FFT-inspired sampling procedure, as a replacement for self-attention, stems from the fact that every linear-combination of sequence embeddings that is generated by contraction with the graph matrix is funneled through the same position-wise feedforward network. An effective flow of information through the network thus requires that some structure be encoded into the embeddings, which allows the feedforward network to disambiguate different elements of the sequence. This is identical to the process used to generate positional embeddings. FFT coefficients naturally provide a powerful schema for sampling linear combinations of vectorized representations in a way that maximizes the distinguishability between different components (because that's like what Fourier transforms do).

WORK IN PROGRESS: For now the FNetAR algorithm exists as (1) further evidence that numerous compounded computations of a structure graph are superfluous for many tasks in natural language understanding (2) a systematic and simplistic method for parameter reduction, applicable to any recurrent Transformer model. Although FNetAR should also be faster than its Transformer-XL counterpart, we are currently working on optimizing the autoregressive Fourier transform and will not be able to comment on the gains in training speed until v1 of this note is released. This updated v1 will also include a comparison of the large models, as well as combinations with the Feedback Transformer, which is likely highly optimized relative to Transformer-XL. There is also reason to believe that FNet may improve the interpretability of the attention-score graphs. Since FNet squeezes the structure-learning ability of standard Transformers into a fewer number of layers, the relationships learned will be fewer and thus each will likely be more meaningful. A cursory exploration of this should also be expected.

Despite the fact that this sampling strategy does not produce an overall transformation that resembles a Fourier transform mathematically, we find these experiments to be useful for thinking about the question of how to optimize the extraction of information using Fourier duality. All evidence indicates that intellectually useful information exists at multiple scales and is coded in both local and non-local correlations. We thus find it plausible that Fourier transforms may be a salient component of systems that efficiently extract both local and non-local information. Subsequently its autoregressive generalizations would be necessary for adaptation to tasks such as time-series prediction and causal inference. Indeed we find it highly likely both that (1) existing architectures such as convolutional networks are already leveraging the equivalence between kernel-convolutions and Fourier transforms (2) there exist additional time invariant or convolutional causal forms that could be used to construct further optimized sampling strategies. 



\clearpage
\newpage
\bibliography{fourier}

\begin{thebibliography}{17}
\providecommand{\natexlab}[1]{#1}
\providecommand{\url}[1]{\texttt{#1}}
\expandafter\ifx\csname urlstyle\endcsname\relax
  \providecommand{\doi}[1]{doi: #1}\else
  \providecommand{\doi}{doi: \begingroup \urlstyle{rm}\Url}\fi

\bibitem[Vaswani et~al.(2017)Vaswani, Shazeer, Parmar, Uszkoreit, Jones, Gomez,
  Kaiser, and Polosukhin]{NIPS2017_3f5ee243}
Ashish Vaswani, Noam Shazeer, Niki Parmar, Jakob Uszkoreit, Llion Jones,
  Aidan~N Gomez, \L~ukasz Kaiser, and Illia Polosukhin.
\newblock Attention is all you need.
\newblock In I.~Guyon, U.~V. Luxburg, S.~Bengio, H.~Wallach, R.~Fergus,
  S.~Vishwanathan, and R.~Garnett, editors, \emph{Advances in Neural
  Information Processing Systems}, volume~30. Curran Associates, Inc., 2017.
\newblock URL
  \url{https://proceedings.neurips.cc/paper/2017/file/3f5ee243547dee91fbd053c1c4a845aa-Paper.pdf}.

\bibitem[Devlin et~al.(2018)Devlin, Chang, Lee, and Toutanova]{devlin2018bert}
Jacob Devlin, Ming-Wei Chang, Kenton Lee, and Kristina Toutanova.
\newblock Bert: Pre-training of deep bidirectional transformers for language
  understanding.
\newblock \emph{arXiv preprint arXiv:1810.04805}, 2018.

\bibitem[Dai et~al.(2019)Dai, Yang, Yang, Carbonell, Le, and
  Salakhutdinov]{dai2019transformer}
Zihang Dai, Zhilin Yang, Yiming Yang, Jaime Carbonell, Quoc~V Le, and Ruslan
  Salakhutdinov.
\newblock Transformer-xl: Attentive language models beyond a fixed-length
  context.
\newblock \emph{arXiv preprint arXiv:1901.02860}, 2019.

\bibitem[Radford et~al.(2018)Radford, Narasimhan, Salimans, and
  Sutskever]{radford2018improving}
Alec Radford, Karthik Narasimhan, Tim Salimans, and Ilya Sutskever.
\newblock Improving language understanding by generative pre-training.
\newblock 2018.

\bibitem[Radford et~al.(2019)Radford, Wu, Child, Luan, Amodei, Sutskever,
  et~al.]{radford2019language}
Alec Radford, Jeffrey Wu, Rewon Child, David Luan, Dario Amodei, Ilya
  Sutskever, et~al.
\newblock Language models are unsupervised multitask learners.
\newblock \emph{OpenAI blog}, 1\penalty0 (8):\penalty0 9, 2019.

\bibitem[Brown et~al.(2020)Brown, Mann, Ryder, Subbiah, Kaplan, Dhariwal,
  Neelakantan, Shyam, Sastry, Askell, et~al.]{brown2020language}
Tom~B Brown, Benjamin Mann, Nick Ryder, Melanie Subbiah, Jared Kaplan, Prafulla
  Dhariwal, Arvind Neelakantan, Pranav Shyam, Girish Sastry, Amanda Askell,
  et~al.
\newblock Language models are few-shot learners.
\newblock \emph{arXiv preprint arXiv:2005.14165}, 2020.

\bibitem[Yang et~al.(2019)Yang, Dai, Yang, Carbonell, Salakhutdinov, and
  Le]{yang2019xlnet}
Zhilin Yang, Zihang Dai, Yiming Yang, Jaime Carbonell, Russ~R Salakhutdinov,
  and Quoc~V Le.
\newblock Xlnet: Generalized autoregressive pretraining for language
  understanding.
\newblock \emph{Advances in neural information processing systems}, 32, 2019.

\bibitem[Wang et~al.(2018)Wang, Girshick, Gupta, and He]{wang2018non}
Xiaolong Wang, Ross Girshick, Abhinav Gupta, and Kaiming He.
\newblock Non-local neural networks.
\newblock In \emph{Proceedings of the IEEE conference on computer vision and
  pattern recognition}, pages 7794--7803, 2018.

\bibitem[Fan et~al.(2020)Fan, Lavril, Grave, Joulin, and
  Sukhbaatar]{fan2020addressing}
Angela Fan, Thibaut Lavril, Edouard Grave, Armand Joulin, and Sainbayar
  Sukhbaatar.
\newblock Addressing some limitations of transformers with feedback memory.
\newblock \emph{arXiv preprint arXiv:2002.09402}, 2020.

\bibitem[Bapna et~al.(2020{\natexlab{a}})Bapna, Arivazhagan, and
  Firat]{bapna2020controlling}
Ankur Bapna, Naveen Arivazhagan, and Orhan Firat.
\newblock Controlling computation versus quality for neural sequence models.
\newblock \emph{arXiv preprint arXiv:2002.07106}, 2020{\natexlab{a}}.

\bibitem[Choromanski et~al.(2020{\natexlab{a}})Choromanski, Likhosherstov,
  Dohan, Song, Gane, Sarlos, Hawkins, Davis, Belanger, Colwell,
  et~al.]{choromanski2020masked}
Krzysztof Choromanski, Valerii Likhosherstov, David Dohan, Xingyou Song,
  Andreea Gane, Tamas Sarlos, Peter Hawkins, Jared Davis, David Belanger, Lucy
  Colwell, et~al.
\newblock Masked language modeling for proteins via linearly scalable
  long-context transformers.
\newblock \emph{arXiv preprint arXiv:2006.03555}, 2020{\natexlab{a}}.

\bibitem[Choromanski et~al.(2020{\natexlab{b}})Choromanski, Likhosherstov,
  Dohan, Song, Gane, Sarlos, Hawkins, Davis, Mohiuddin, Kaiser,
  et~al.]{choromanski2020rethinking}
Krzysztof Choromanski, Valerii Likhosherstov, David Dohan, Xingyou Song,
  Andreea Gane, Tamas Sarlos, Peter Hawkins, Jared Davis, Afroz Mohiuddin,
  Lukasz Kaiser, et~al.
\newblock Rethinking attention with performers.
\newblock \emph{arXiv preprint arXiv:2009.14794}, 2020{\natexlab{b}}.

\bibitem[Bapna et~al.(2020{\natexlab{b}})Bapna, Arivazhagan, and
  Firat]{DBLP:journals/corr/abs-2002-07106}
Ankur Bapna, Naveen Arivazhagan, and Orhan Firat.
\newblock Controlling computation versus quality for neural sequence models.
\newblock \emph{CoRR}, abs/2002.07106, 2020{\natexlab{b}}.
\newblock URL \url{https://arxiv.org/abs/2002.07106}.

\bibitem[Choromanski et~al.(2020{\natexlab{c}})Choromanski, Likhosherstov,
  Dohan, Song, Davis, Sarl{\'{o}}s, Belanger, Colwell, and
  Weller]{DBLP:journals/corr/abs-2006-03555}
Krzysztof Choromanski, Valerii Likhosherstov, David Dohan, Xingyou Song, Jared
  Davis, Tam{\'{a}}s Sarl{\'{o}}s, David Belanger, Lucy~J. Colwell, and Adrian
  Weller.
\newblock Masked language modeling for proteins via linearly scalable
  long-context transformers.
\newblock \emph{CoRR}, abs/2006.03555, 2020{\natexlab{c}}.
\newblock URL \url{https://arxiv.org/abs/2006.03555}.

\bibitem[Choromanski et~al.(2020{\natexlab{d}})Choromanski, Likhosherstov,
  Dohan, Song, Gane, Sarl{\'{o}}s, Hawkins, Davis, Mohiuddin, Kaiser, Belanger,
  Colwell, and Weller]{DBLP:journals/corr/abs-2009-14794}
Krzysztof Choromanski, Valerii Likhosherstov, David Dohan, Xingyou Song,
  Andreea Gane, Tam{\'{a}}s Sarl{\'{o}}s, Peter Hawkins, Jared Davis, Afroz
  Mohiuddin, Lukasz Kaiser, David Belanger, Lucy~J. Colwell, and Adrian Weller.
\newblock Rethinking attention with performers.
\newblock \emph{CoRR}, abs/2009.14794, 2020{\natexlab{d}}.
\newblock URL \url{https://arxiv.org/abs/2009.14794}.

\bibitem[Lee-Thorp et~al.(2021)Lee-Thorp, Ainslie, Eckstein, and
  Ontanon]{lee2021fnet}
James Lee-Thorp, Joshua Ainslie, Ilya Eckstein, and Santiago Ontanon.
\newblock Fnet: Mixing tokens with fourier transforms.
\newblock \emph{arXiv preprint arXiv:2105.03824}, 2021.

\bibitem[Buchholz and Jug(2021)]{DBLP:journals/corr/abs-2104-02555}
Tim{-}Oliver Buchholz and Florian Jug.
\newblock Fourier image transformer.
\newblock \emph{CoRR}, abs/2104.02555, 2021.
\newblock URL \url{https://arxiv.org/abs/2104.02555}.

\end{thebibliography}

\end{document}